\documentclass[11pt]{article}

\usepackage[preprint]{acl}
\usepackage{algorithm}
\usepackage{algorithmic}
\usepackage{amsmath}
\usepackage{amssymb}
\usepackage{amsthm}

\theoremstyle{definition}
\newtheorem{definition}{Definition}
\usepackage{booktabs}
\usepackage{xspace}
\usepackage{multirow}
\usepackage{array}
\usepackage{tabularx}

\usepackage{times}
\usepackage{latexsym}

\usepackage[T1]{fontenc}
\usepackage[utf8]{inputenc}

\usepackage{microtype}

\usepackage{inconsolata}

\usepackage{graphicx}

\newcommand{\method}{\textsc{HarnessLens}\xspace}
\newcommand{\secref}[1]{Section~\ref{#1}}
\newcommand{\vheadl}[1]{%
  \begin{tabular}[c]{@{}l@{}}#1\end{tabular}}
\newcommand{\vheadc}[1]{%
  \begin{tabular}[c]{@{}c@{}}#1\end{tabular}}

\title{Verify Smarter, Evolve Further: Efficient Harness Evolution\\ through Behavior-Aware Verification}

\author{
 \textbf{Jinghan Xu\textsuperscript{1,2}},
 \textbf{Yikai Zhang\textsuperscript{2,3}},
 \textbf{Aili Chen\textsuperscript{2,3}},
 \textbf{Weiyuan Li\textsuperscript{1,2}},
 \textbf{Jiaqing Liang\textsuperscript{1,2}},
 \textbf{Deqing Yang\textsuperscript{1,2}\thanks{Corresponding author.}}
\\
 \textsuperscript{1}School of Data Science, Fudan University,\\
 \textsuperscript{2}Shanghai Key Laboratory of Data Science, \\
 \textsuperscript{3}College of Computer Science and Artificial Intelligence, Fudan University
\\
 \texttt{{jhxu25@m.fudan.edu.cn, yangdeqing@fudan.edu.cn}
 }
}

\begin{document}
\maketitle
\begin{abstract}
Agent harnesses shape how language-model agents use instructions, tools, and runtime components, but adapting these harnesses requires costly verification. 
Existing propose-and-verify methods typically score every candidate on a fixed task set, wasting rollouts on unrelated behaviors and allowing aggregate scores to obscure specific regressions.
We introduce \method, a budget-aware framework for automated harness evolution. \method jointly explores the task space and user-configurable components, derives candidate modifications from execution trajectories, and selectively verifies each candidate on behavior-relevant tasks using an attributable-evidence gate. 
Across three agent harnesses and four benchmarks, \method improves average held-out performance by 7.6--13.6\% while consuming substantially less evaluation budget than competing baselines. 
These results demonstrate that behavior-aware verification with explicit attribution enables more reliable and sample-efficient harness evolution under constrained interaction budgets.
Our code is available at \url{https://github.com/jhxu5214/HarnessLens}.
\end{abstract}

\section{Introduction}

\begin{figure}[t]
\centering
% Reduce the figure size so that it is slightly narrower than the column.
% Don't use precise values for figure width.
% This setup will avoid overfull boxes.
\includegraphics[width=0.95\columnwidth]{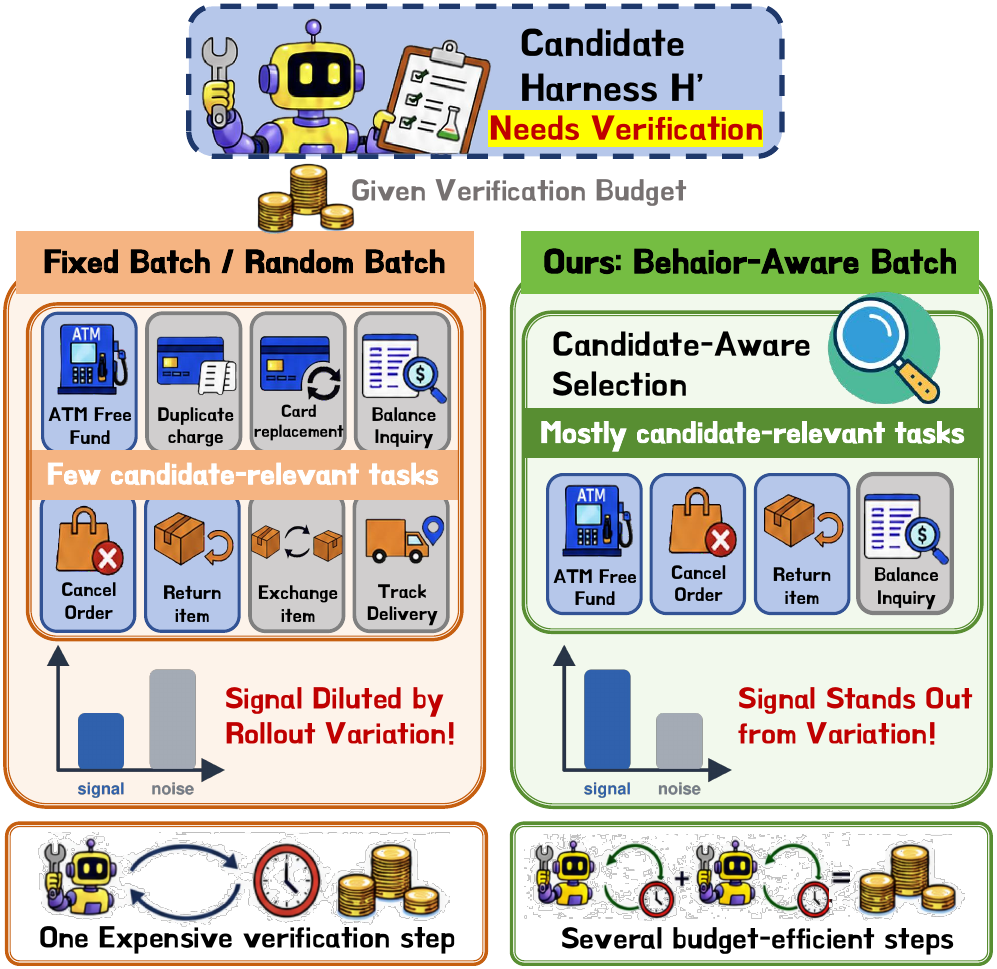}
% \caption{\sout{Motivation for \method.}
% A batch selected without considering the current modification may spend the verification budget on tasks that do not exercise the intended behavior, diluting the candidate's effect.
% \zyk{Existing works (left) xxx}
% \method instead selects each verification batch for the behaviors the current modification is expected to affect.}
\caption{\textbf{Comparison of verification paradigms:} Fixed or random batches dilute candidate-specific signals, while our behavior-aware batches enable clearer and more efficient verification.}
\label{fig:motivation}
\end{figure}

An agent harness determines how a language-model agent perceives tasks, uses tools, and acts in an environment through diverse components such as instructions, skills, tool descriptions, memory, permissions, and agent roles~\cite{lin2026ahe,chen2026harnessfix,ning2026code}.
It substantially shapes how large language models (LLMs) execute tasks and significantly affects their performance~\cite{lee2026metaharness,lin2026ahe}.
However, building an effective harness is usually nontrivial and requires substantial manual effort, while a configuration developed for one model or environment may not be transferable.
This raises a critical question: \textbf{can an agent harness autonomously and continually evolve from the interaction evidence generated during task execution?}

To solve this problem, existing methods generally follow a \emph{propose-and-verify} paradigm: they propose candidate modifications based on previous interactions and verify them through additional task rollouts. This paradigm is commonly applied to prompt and program optimization~\cite{yang2024opro,opsahlong2024mipro,yuksekgonul2024textgrad} and is increasingly applied to richer harness components~\cite{lou2026autoharness,hao2026recreate,lee2026metaharness,zhang2026selfharness,lin2026ahe,chen2026harnessfix,huang2026memoharness}. Across these methods, verification typically relies on fixed task sets~\cite{lin2026ahe,chen2026harnessfix}, randomly sampled minibatches~\cite{agrawal2026gepa}, or task subsets preselected before evolution~\cite{pan2026dark}, with some methods further adopting staged screening~\cite{luo2026gsme}.

However, these methods suffer from a fundamental limitation: they use the same evaluation tasks for different modifications, even when these modifications target different behaviors.
As illustrated in Figure~\ref{fig:motivation}, many tasks in such sets may be unrelated to the intended effect of a modification, causing unnecessary rollouts and diluting the modification signal in aggregate metrics.
Moreover, verification results may fail to reveal whether the intended behavior has actually emerged when the selected tasks do not cover the affected behaviors.
Although recent methods improve trajectory diagnosis~\cite{lin2026ahe,chen2026harnessfix} or reduce verification cost through task sampling~\cite{agrawal2026gepa}, preselected task coresets~\cite{pan2026dark}, or staged screening~\cite{luo2026gsme}, a key limitation persists: verification tasks are still not adapted to the behavioral effect of each modification, and evaluation evidence is not systematically reused to guide subsequent evolution.

To address this gap, we introduce \method, a budget-aware framework for autonomous harness evolution with behavior-aware verification.
It selects verification tasks based on supporting trajectories, task patterns, affected components, and regression risks, and compares candidate trajectories to attribute behavioral changes and preserve existing capabilities.
\method contains three stages: \textbf{Context Exploration} characterizes tasks and user-configurable components, \textbf{Trajectory Diagnosis} extracts reusable experiences and deficiencies from rollouts, and \textbf{Harness Evolution} constructs and verifies modifications using the collected evidence.
Each rollout therefore evaluates current candidates while contributing evidence for future evolution.

We evaluate \method on three agent harnesses across four benchmarks: $\tau^3$-bench Banking Knowledge~\cite{shi2026tau}, $\tau^2$-bench Retail~\cite{barres2025tau2}, Terminal-Bench 2.0~\cite{merrill2026terminalbench}, and the Challenging subset of BIRD Mini-Dev~\cite{li2024can}.
Under the given budgets, \method improves initial harnesses and outperforms fixed-set verification methods, yielding average performance improvements of 7.6--13.6\% across harnesses with a substantially smaller joint rollout-and-analysis budget.

Our contributions are as follows:
\begin{itemize}

\item We propose a behavior-aware approach to harness self-evolution that selects verification tasks and allocates rollouts for each modification, instead of evaluating all modifications on the same task set.

\item We introduce \method, a budget-aware and harness-independent framework that discovers user-configurable components and their update mechanisms at runtime, diagnoses interaction trajectories, and iteratively proposes, verifies, and reviews harness modifications.

\item We evaluate \method across three agent harnesses and four benchmarks. It achieves better test performance than existing methods with substantially fewer rollouts, improving success rates by up to 13.6\% on OpenCode, 7.6\% on Codex, and 9.2\% on Pi.

\end{itemize}

\begin{figure*}[t]
\centering
\includegraphics[width=0.96\textwidth]{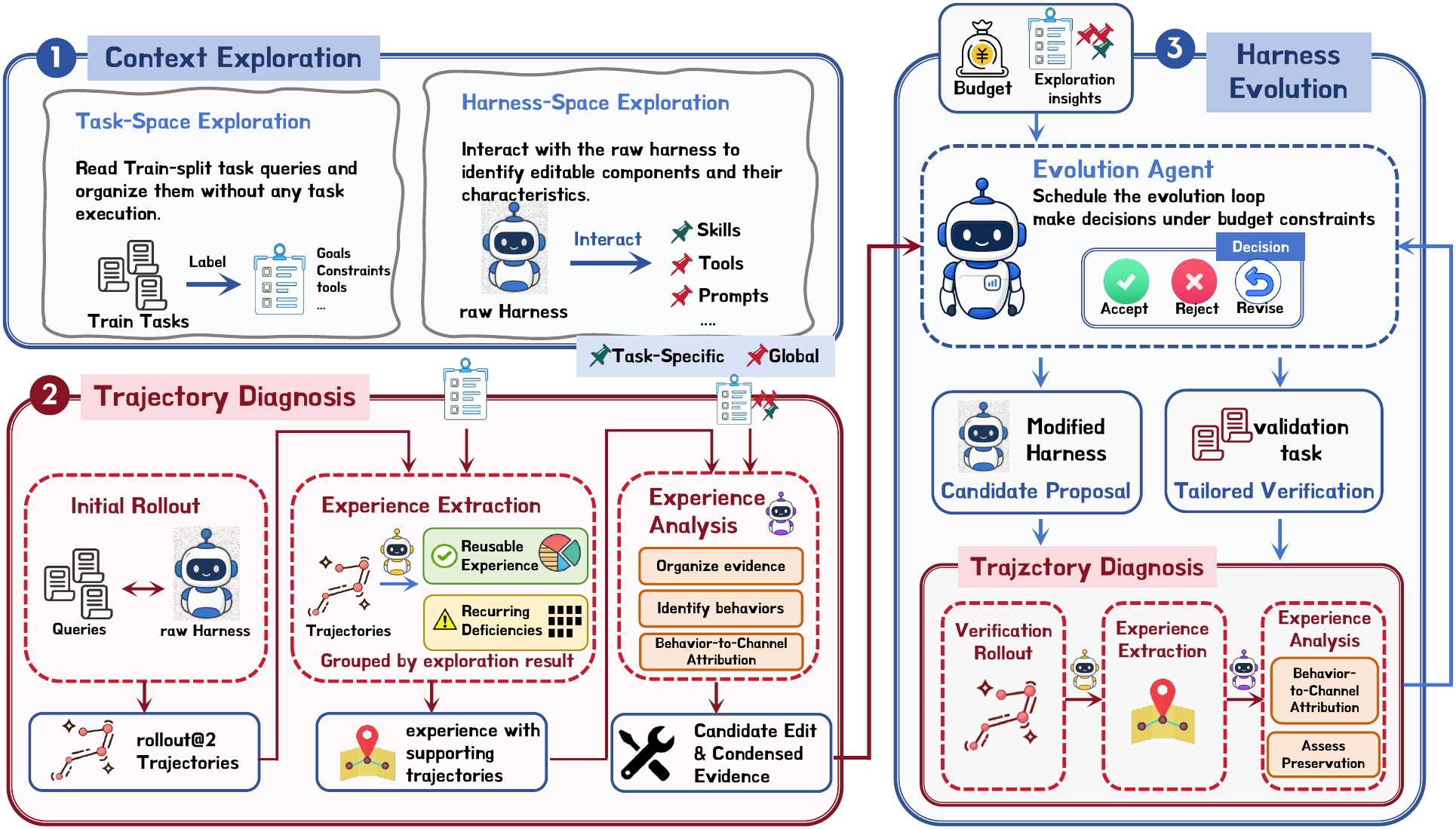}
\caption{Overview of \method. 
Context Exploration characterizes available tasks and identifies user-configurable components. Trajectory Diagnosis extracts and analyzes rollout evidence and is reused within Harness Evolution, which constructs and verifies modifications. The entire process is constrained by a fixed interaction budget.}
\label{fig:overview}
\end{figure*}

\section{Related Work}
\paragraph{Self-improving agents and harness evolution.}
Self-improving agents use interaction feedback to revise the scaffolds that shape their behavior.
Earlier methods operate on relatively lightweight structures, improving prompts~\cite{agrawal2026gepa}, workflow graphs~\cite{zhang2025aflow}, prompt-level harness specifications~\cite{lee2026rhi}, or domain scaffolds~\cite{hao2026recreate}.
Harness evolution expands this scope: some methods synthesize or rewrite harness programs~\cite{lou2026autoharness,lee2026metaharness}, while others update user-configurable components such as prompts, tools, skills, memory, middleware, and orchestration~\cite{lin2026ahe,zhang2026selfharness,huang2026memoharness}, or accumulate trajectory-derived procedural state~\cite{du2026livingharness}.
% Recent harness evolution methods broaden the scope of harness modification to include executable agent logic and user-configurable components, while related work supports behavior-to-source localization for harness modification~\cite{wang2026handbook}.
% Some synthesize or rewrite harness programs~\cite{lou2026autoharness,lee2026metaharness}, while others update combinations of prompts, tools, skills, memory, middleware, orchestration, and trajectory-derived procedural state~\cite{lin2026ahe,zhang2026selfharness,huang2026memoharness,du2026livingharness}.
Harness evolution has also been studied for nonstationary task streams~\cite{liu2026adaptiveharness}, long-running environments~\cite{karten2026continualharness}, and test-time adaptation without labels~\cite{nie2026tthe}.
However, candidate-based approaches typically use fixed or shared verification tasks across modifications, wasting rollouts on unaffected behaviors and diluting modification-specific evidence.
\method instead selects informative verification tasks for each modification.

\paragraph{Harness verification and rollout allocation.}
Existing methods use verification rollouts to decide whether to accept candidate modifications.
Most evaluate all candidates on a shared, predefined task collection, including fixed or held-out sets~\cite{lee2026metaharness,zhang2026selfharness,huang2026memoharness} and current evaluation batches~\cite{nie2026tthe}.
Other approaches narrow verification to predefined flaw- or failure-related tasks~\cite{chen2026harnessfix,cai2026moss} or reduce cost using a globally preselected coreset~\cite{pan2026dark}.
% Harness Scaling~\cite{wang2026rethinkingharness} emphasizes matched inference budgets and held-out evaluation, but focuses on final comparison rather than in-loop rollout allocation. 
In contrast, \method adapts verification tasks and rollout allocation to each modification's intended behavior, affected components, regression risks, and accumulated evidence, reducing irrelevant rollouts while preserving modification-specific signals.

\section{Problem Formulation}
\label{sec:problem-formulation}

We first define the optimization scope of harness evolution, including the user-configurable components of a harness framework, the resulting harness space, and the modifications used to generate candidate harnesses.
We then formulate harness evolution as a budget-constrained optimization problem.

\subsection{Definitions}
\label{sec:definitions}

\begin{definition}[\textbf{Components and Harness}]
A harness framework \(\mathcal{F}\) exposes a set of user-configurable components,
\[
\mathcal{C}_{\mathcal{F}}=\{c_1,\ldots,c_m\}.
\]

These components may include instructions, skills, prompt templates, tools and integrations, agent roles, and runtime extensions exposed through predefined interfaces.
Each component \(c\in\mathcal{C}_{\mathcal{F}}\) has a configuration space \(\mathcal{X}_c\), whose elements specify the content and instances configured for that component.
A concrete harness is represented as
\[
\mathcal{H}
=
\bigl(\mathcal{F},\mathbf{h}\bigr),
\quad
\mathbf{h}=(h_c)_{c\in\mathcal{C}_{\mathcal{F}}},
\quad
h_c\in\mathcal{X}_c.
\]
\end{definition}

For a fixed framework \(\mathcal{F}\), we define the admissible harness space as
\[
\mathbb{H}_{\mathcal{F}}
\subseteq
\{\mathcal{F}\}
\times
\prod_{c\in\mathcal{C}_{\mathcal{F}}}\mathcal{X}_c.
\]

Harnesses in this space may differ in the content and instances configured through the user-configurable components, while sharing the same component types, execution mechanisms, and underlying framework.

\begin{definition}[\textbf{Agent}]
An agent is a base language model operating within a concrete harness:
\[
\mathcal{A}_{\theta,\mathcal{H}}
=
(M_\theta,\mathcal{H}),
\]
where \(M_\theta\) is a base language model parameterized by \(\theta\) and \(\mathcal{H}\in\mathbb{H}_{\mathcal{F}}\).
\end{definition}

\begin{definition}[\textbf{Modification and Candidate}]
Under a fixed framework \(\mathcal{F}\), an admissible harness modification operates on the joint component configuration:
\[
\delta:
\prod_{c\in\mathcal{C}_{\mathcal{F}}}\mathcal{X}_c
\rightarrow
\prod_{c\in\mathcal{C}_{\mathcal{F}}}\mathcal{X}_c.
\]

Given a harness \(\mathcal{H}=(\mathcal{F},\mathbf{h})\), applying \(\delta\) produces a candidate harness
\[
\mathcal{H}'
=
\bigl(\mathcal{F},\delta(\mathbf{h})\bigr)
\in\mathbb{H}_{\mathcal{F}}.
\]

Thus, the modification changes the configuration of one or more user-configurable components while leaving \(\mathcal{F}\) unchanged.
\end{definition}

Throughout this work, we consider harness evolution under a fixed base model and a fixed harness framework.
Thus, \(\theta\) and \(\mathcal{F}\) remain unchanged, while evolution operates only over the admissible component configurations in \(\mathbb{H}_{\mathcal{F}}\).

\subsection{Optimization Objective}
\label{sec:optimization-objective}

For a task \(x\sim\mathcal{D}\), the agent \(\mathcal{A}_{\theta,\mathcal{H}}\) induces a trajectory
\[
\tau=(o_0,a_0,o_1,a_1,\ldots,o_T),
\;
\tau\sim p(\tau\mid x;M_\theta,\mathcal{H}),
\]
where \(o_t\) and \(a_t\) denote the observation and action at step \(t\), respectively.
Each trajectory receives a binary reward \(R(x,\tau)\in\{0,1\}\), indicating whether the task is successfully completed.
We refer to a \emph{behavior} as a recurring, model-visible pattern within such trajectories: which actions the agent takes, in what order, and under which conditions.
Behaviors are identified from trajectories rather than from rewards, so trajectories receiving the same reward may still differ in behavior.

Let \(\mathcal{T}_{\mathrm{train}}\) and \(\mathcal{T}_{\mathrm{test}}\) be disjoint task sets sampled from the target task distribution \(\mathcal{D}\).
Only \(\mathcal{T}_{\mathrm{train}}\) is accessible during harness evolution, whereas \(\mathcal{T}_{\mathrm{test}}\) remains held out for final evaluation.

Starting from an initial harness \(\mathcal{H}_0\in\mathbb{H}_{\mathcal{F}}\), our objective is to find an evolved harness that maximizes the expected task success rate under a fixed interaction budget:
\begin{equation*}
\begin{aligned}
\mathcal{H}^{*}
&=
\arg\max_{\mathcal{H}\in\mathbb{H}_{\mathcal{F}}}
\mathbb{E}_{\substack{
x\sim\mathcal{D},\,
\tau\sim p(\tau\mid x;M_\theta,\mathcal{H})
}}
\left[R(x,\tau)\right], \\
&\qquad
\text{s.t.}\quad
C(\mathcal{H}_0\rightsquigarrow\mathcal{H})\leq B,
\end{aligned}
\end{equation*}
where $B$ denotes the interaction budget and \(C(\mathcal{H}_0\rightsquigarrow\mathcal{H})\) denotes the number of LLM sessions and task trials consumed along the evolution path from \(\mathcal{H}_0\) to \(\mathcal{H}\), including the initial rollout, Context Exploration, Trajectory Diagnosis, and Harness Evolution.
Here, an LLM session refers to one complete multi-turn execution of a model-based role.
Since \(\mathcal{T}_{\mathrm{train}}\) participates in harness evolution, generalization performance is measured by the pass rate on the held-out set \(\mathcal{T}_{\mathrm{test}}\).

\section{Method}
\label{sec:method}
% Under the budgeted objective in \secref{sec:optimization-objective}, harness evolution must decide both how to modify the harness and where to spend limited interactions.
% Given \(\mathcal{H}_0=(\mathcal{F},\mathbf{h}_0)\), \(\mathcal{T}_{\mathrm{train}}\), and \(B\), \method improves \(\mathbf{h}_0\) while holding \(\mathcal{F}\) fixed.
% An \emph{evolution agent}, distinct from \(\mathcal{A}_{\theta,\mathcal{H}}\), selects proposals and verification tasks and decides whether to accept, reject, or revise each modification, or to return a fresh hypothesis to the proposal pool, using diagnostic reports, regression risk, and the remaining budget.
% Every model-based role, including this agent, is itself instantiated on the framework \(\mathcal{F}\) under evolution, so the method requires no agent scaffolding outside the target harness.
% Each model-based invocation and task trial consumes one unit of \(B\).
% A deterministic, non-model \emph{controller} enforces budget accounting, isolation, trial and batch limits, rollout reuse, and promotion gates.
% Implementation details, including role boundaries, editable components, and verification, appear in Appendix~\ref{sec:appendix-procedure}, \ref{sec:appendix-harness-spaces}, and \ref{sec:appendix-behavior-verification}.

Under the budgeted objective in \secref{sec:optimization-objective}, \method evolves the configurable state \(\mathbf{h}_0\) within the fixed harness framework \(\mathcal{F}\) under interaction budget \(B\). 
A deterministic controller coordinates the procedure, tracks budget consumption, isolates candidate executions, manages rollout reuse, and enforces update criteria. 
Further implementation details appear in Appendix~\ref{sec:appendix-procedure}, \ref{sec:appendix-harness-spaces}, and \ref{sec:appendix-behavior-verification}.

As illustrated in Figure~\ref{fig:overview}, \method contains three components: \textbf{Context Exploration} (\secref{sec:context-exploration}) characterizes tasks and editable components; \textbf{Trajectory Diagnosis} (\secref{sec:trajectory-diagnosis}) processes trajectories from initial and verification rollouts; and \textbf{Harness Evolution} (\secref{sec:harness-evolution}) uses its outputs to select proposals, construct candidate harnesses, and verify them before updating the current harness.

\subsection{Context Exploration}
\label{sec:context-exploration}

Context Exploration provides the task organization and editable-component information needed for subsequent behavior-aware verification.
\emph{Task-Space Exploration} characterizes task organization from available task context, while \emph{Harness-Space Exploration} identifies the user-configurable components of \(\mathcal{F}\) and their editable scope.
Their outputs ground Trajectory Diagnosis and focus later rollouts on relevant behaviors and regressions.

\paragraph{Task-Space Exploration.}
To support behavior-aware verification, \textit{Task-Space Exploration} organizes \(\mathcal{T}_{\mathrm{train}}\) according to task goals.
For each task, the module inspects only the query, environment instructions and policies, and tool and parameter descriptions.
It assigns tasks to groups based on their primary user goals and records additional goals when a task spans multiple groups.
This grouping provides a compact view of recurring task patterns and overlaps across goals.
The resulting task groups guide verification-task selection and regression checks but are not incorporated into \(\mathbf{h}\).
At this stage, \textit{Task-Space Exploration} executes no tasks and has no access to trajectories, task outcomes, or \(\mathcal{T}_{\mathrm{test}}\).

\paragraph{Harness-Space Exploration.}
To identify the components available for evolution, \textit{Harness-Space Exploration} examines the configuration, documentation, and runtime behavior of \(\mathcal{F}\).
For each user-configurable component \(c\in\mathcal{C}_{\mathcal{F}}\), the module records how it is exposed to the agent, where its effects apply, how it can be updated, and what behaviors may be affected by a change.
Only components that can be reliably identified and updated are retained for subsequent evolution.
These component descriptions later help connect diagnosed behaviors to candidate modifications and potential regressions.

\subsection{Trajectory Diagnosis}
\label{sec:trajectory-diagnosis}

Building on \textit{Context Exploration}, \textit{Trajectory Diagnosis} converts trajectories into behavioral evidence through \emph{Experience Extraction} and \emph{Experience Analysis}.
It is applied to the \emph{initial rollout} under \(\mathcal{H}_0\) and to each \emph{verification rollout} during \textit{Harness Evolution}.

\paragraph{Experience Extraction.}
Task rewards alone do not reveal which behaviors should be preserved or improved.
The Experience module summarizes trajectories under analysis into \emph{reusable experiences} and \emph{recurring deficiencies}.
Each extracted item is linked to the trajectories that support it.
To organize this evidence, recurring behaviors are grouped across trajectories, while distinct successful strategies are kept separate.
The resulting evidence is passed to \emph{Experience Analysis} for further assessment.

\paragraph{Experience Analysis.}
To derive candidate modifications from this evidence, the Analyzer combines the extracted experiences and deficiencies with the task groups and identified components from \textit{Context Exploration}, producing a set of modification proposals.
Each proposal specifies the targeted behavior, the trajectories that support it, and the user-configurable components that can affect that behavior.
The Analyzer then checks whether the supporting trajectories provide sufficient evidence for the proposed modification and removes proposals that lack such support.
The resulting proposals and their supporting evidence are passed to \textit{Harness Evolution}, while the extracted experiences remain available for later regression checks.

\subsection{Harness Evolution}
\label{sec:harness-evolution}
Using the proposals and supporting evidence from \textit{Trajectory Diagnosis}, \textit{Harness Evolution} iteratively improves the current confirmed harness \(\mathcal{H}_j=(\mathcal{F},\mathbf{h}_j)\).
Harness Evolution is carried out by an evolution agent, a model-based role that uses the same harness framework \(\mathcal{F}\) as the target agent \(\mathcal{A}_{\theta,\mathcal{H}}\) but operates independently from it.
At each iteration, the evolution agent selects a proposal in \emph{Candidate Proposal}, chooses its verification tasks in \emph{Behavior-Aware Verification}, and decides whether the resulting candidate should replace \(\mathcal{H}_j\) in \emph{Harness Review and Update}.

\paragraph{Candidate Proposal.}

At each iteration, the evolution agent considers one proposal from \textit{Trajectory Diagnosis}.
The proposal is selected based on its supporting evidence, previous attempts, potential regressions, and the remaining budget.
The corresponding modification is applied to a copy of the current confirmed harness using the update mechanism identified for the affected component.
Let \(\delta_j\) denote this modification.
Applying \(\delta_j\) to \(\mathbf{h}_j\) produces the candidate harness
\[
\widetilde{\mathcal{H}}_j
=
\bigl(\mathcal{F},\delta_j(\mathbf{h}_j)\bigr)
\in \mathbb{H}_{\mathcal{F}}.
\]

Each new candidate is derived from the current confirmed harness \(\mathcal{H}_j\), allowing accepted modifications to accumulate across iterations.
Before verification, a lightweight runtime check confirms that the modified component has been successfully applied.
Candidates that fail this check do not proceed to verification.

\paragraph{Behavior-Aware Verification.}

A fixed verification batch may contain few tasks relevant to the behavior targeted by a proposal.
To focus verification on that behavior, the evolution agent first selects tasks in \(\mathcal{T}_{\mathrm{train}}\) linked to the proposal's supporting trajectories and then selects additional tasks with related goals, constraints, or tool requirements.
It also includes tasks that can reveal potential regressions, with broader modifications receiving coverage across more task groups.
Each verification batch contains at least five distinct tasks, while the controller fixes the number of trials and ensures that the required verification can be completed within the remaining budget.

The selected tasks are evaluated under both \(\mathcal{H}_j\) and \(\widetilde{\mathcal{H}}_j\) using matched trial conditions.
\textit{Trajectory Diagnosis} is then applied to the resulting trajectories to determine whether \(\delta_j\) improves the targeted behavior and introduces any regressions.

\paragraph{Harness Review and Update.}

The evolution agent reviews the behavioral evidence and rollout metrics to determine whether the observed improvement is supported by the trajectories and whether the candidate introduces regressions.

Candidates with supported improvements and no observed regression undergo an additional confirmation on a new batch constructed by the controller. 
The batch retains at most two tasks that require further confirmation and fills the remaining positions with previously unused tasks from \(\mathcal{T}_{\mathrm{train}}\), prioritizing task groups not covered in the initial verification.

Both \(\mathcal{H}_j\) and \(\widetilde{\mathcal{H}}_j\) are evaluated on this batch to check whether the observed improvement persists and whether the modification affects other tasks.
The final decision considers the behavioral evidence together with the primary metric on the confirmation batch; improvement in the primary metric alone is insufficient.

The harness is updated according to
$$
\mathcal{H}_{j+1}
=
\begin{cases}
\widetilde{\mathcal{H}}_j,
& \text{if the candidate is accepted},\\
\mathcal{H}_j,
& \text{otherwise}.
\end{cases}
$$

If the diagnosis identifies a specific issue with \(\delta_j\) that can be addressed, the evolution agent may adjust the proposal and verify the resulting candidate again.
New issues identified during verification may also lead to new proposals.

Evolution ends when no sufficiently supported proposal remains or the remaining budget cannot support another verification cycle.

\section{Experiments}
\label{sec:experiments}
\subsection{Experiment Setup}
\label{sec:experiment-setup}

\paragraph{Benchmarks and Splits.}
We evaluate on four environments requiring diverse agent behaviors: \(\tau^2\)-bench Retail~\cite{barres2025tau2}, \(\tau^3\)-bench Banking Knowledge~\cite{shi2026tau}, Terminal-Bench 2.0~\cite{merrill2026terminalbench}, and the Challenging subset of BIRD Mini-Dev~\cite{li2024can}.
For each benchmark, we randomly sample 30 tasks for TRAIN; TEST uses the official split where available and otherwise all remaining tasks.
The splits are disjoint, and all information from TEST, including queries, trajectories, and evaluation feedback, is inaccessible throughout harness evolution.
Appendix~\ref{sec:appendix-experimental-config} provides details on the data splits and the full benchmark configuration.

\paragraph{Agent Harnesses and Model.}
We conduct experiments on the following harnesses: OpenCode~\cite{opencode2026} v1.17.13, Codex CLI~\cite{openai2026codex} v0.144.4, and Pi Coding Agent~\cite{zechner2025pi} v0.80.10.
All LLM agent and evolution roles use \texttt{deepseek-v4-flash-preview}~\cite{deepseek2026v4}.
To ensure fair evaluation, we disable external information retrieval tools (e.g., web search) and enforce permission controls to prevent answer leakage or unauthorized tool usage.

\paragraph{Baselines.}
We compare the initial, unmodified harness \(\mathcal{H}_0\) with three trajectory-based harness-evolution methods: Self-Harness~\cite{zhang2026selfharness}, Meta-Harness~\cite{lee2026metaharness}, and HarnessFix~\cite{chen2026harnessfix}
\footnote{We follow each baseline's original evolution-budget configuration.
All methods use the same 30 TRAIN tasks within each harness--benchmark setting, while some baselines further partition them internally for validation.}
.
All methods start from \(\mathcal{H}_0\) and share the same benchmark runtime and verifier.
At their configured maxima, Self-Harness, Meta-Harness, and HarnessFix consume 4{,}800, 660, and 300 TRAIN rollouts, respectively, whereas \method is capped at 200 total units, including both task rollouts and LLM sessions.
Appendix~\ref{sec:appendix-baselines} provides the detailed protocols and the calculation of these configured budget maxima.

\subsection{Main Results}

\begin{table*}[t]
\centering
\small
\setlength{\tabcolsep}{1mm}

\begin{tabular}{llccccc}
\toprule

\multicolumn{1}{c}{%
  \vheadc{\textbf{Harness}}}
&
\multicolumn{1}{c}{%
  \vheadc{\textbf{Method}}}
&
\vheadc{\textbf{$\tau^2$-bench}\\
        \textbf{Retail}}
&
\vheadc{\textbf{$\tau^3$-bench}\\
        \textbf{Banking}}
&
\vheadc{\textbf{Terminal-}\\
        \textbf{Bench 2.0}}
&
\vheadc{\textbf{BIRD Mini-Dev}\\
        \textbf{(Challenging)}}
&
\vheadc{\textbf{AVG.}}
\\

\midrule

\multirow[c]{5}{*}{OpenCode}
& \(\mathcal{H}_0\)
& 75.00
& 20.90
& 33.90
& 37.50
& 41.83\\

& Self-Harness
& 80.00
& 10.45
& 32.20
& 37.50
& 40.04\\

& Meta-Harness
& 72.50
& 13.43
& 33.90
& 41.67
& 40.38\\

& HarnessFix
& 80.00
& 22.39
& \textbf{35.59}
& 40.28
& 44.57\\

& \textbf{HarnessLens (Ours)}
& \textbf{85.00}
& \textbf{25.37}
& 33.90
& \textbf{45.83}
& \textbf{47.53}\\

\midrule

\multirow[c]{5}{*}{Codex}
& \(\mathcal{H}_0\)
& \textbf{80.00}
& 13.43
& 32.84
& 37.50
& 40.94\\

& Self-Harness
& \textbf{80.00}
& 13.43
& \textbf{37.29}
& 34.72
& 41.36\\

& Meta-Harness
& 72.50
& \textbf{16.42}
& 30.51
& 37.50
& 39.23\\

& HarnessFix
& \textbf{80.00}
& 13.43
& 35.59
& 37.50
& 41.63\\

& \textbf{HarnessLens (Ours)}
& \textbf{80.00}
& 13.43
& 35.59
& \textbf{47.22}
& \textbf{44.06}\\

\midrule

\multirow[c]{5}{*}{Pi}
& \(\mathcal{H}_0\)
& \textbf{85.00}
& 19.40
& \textbf{37.29}
& 40.28
& 45.49\\

& Self-Harness
& \textbf{85.00}
& 20.90
& 28.81
& 33.33
& 42.01\\

& Meta-Harness
& 80.00
& 16.42
& 33.90
& \textbf{48.61} 
& 44.73\\

& HarnessFix
& \textbf{85.00}
& 26.87
& 35.59
& 44.44 
& 47.98\\

& \textbf{HarnessLens (Ours)}
& \textbf{85.00}
& \textbf{33.33}
& \textbf{37.29}
& 43.06 
& \textbf{49.67}\\

\bottomrule
\end{tabular}

\caption{Performance of different harness-evolution methods across agent harnesses and target benchmarks. Results are reported as pass@1 (\%) on the held-out TEST split.
The best result in each harness--benchmark setting is shown in bold. }
\label{tab:main-results}
\end{table*}

\paragraph{\method performs best under the smallest budget.}
\method attains the best or tied-best pass rate in eight of the twelve harness--benchmark pairs in Table~\ref{tab:main-results}, while using the smallest configured budget of any evolving method: two-thirds of HarnessFix's budget and one twenty-fourth of Self-Harness's budget (Table~\ref{tab:budget}).
The gains over \(\mathcal{H}_0\) are smallest on Retail, where \(\mathcal{H}_0\) is already at least 75\% for every harness, and largest on Banking and BIRD.
The budget gap is understated because \method's budget counts both LLM sessions and task rollouts, whereas the baseline figures count rollouts alone (Appendix~\ref{sec:appendix-baselines}).

\paragraph{Harness evolution is stable and effective.}
\method never falls below \(\mathcal{H}_0\), and its worst outcome is an exact tie.
Such a tie reflects an unchanged harness rather than a small performance difference: when no proposal accumulates attributable improvement without an attributable regression, the review conditions of \secref{sec:harness-evolution} are never satisfied and the run returns \(\mathcal{H}_0\).
Outcomes thus have only two modes: no edit or an effective edit. Together, Self-Harness and Meta-Harness fall below \(\mathcal{H}_0\) in half of their 24 harness--benchmark pairs and lose about ten points at worst.
Behavior-aware verification therefore acts as a filter rather than a search accelerator: an uninformative iteration results in a tie rather than a regression.

\paragraph{More verification rollouts do not imply better harnesses.}
Performance does not increase with verification budget: Self-Harness uses the most rollouts but does not achieve the strongest performance, whereas HarnessFix achieves the strongest baseline results with the smallest budget.
The key difference is how candidates are accepted.
The baselines rely on aggregate pass rates over fixed task splits, where modification-specific gains can be obscured by unrelated tasks and trial noise.
This may allow chance improvements and undetected regressions to accumulate across iterations and become apparent only on TEST.
More candidates therefore create more opportunities to accept harmful edits rather than reliably improving the harness.

% \paragraph{More verification rollouts do not imply better harnesses.}
% Budget size does not order the methods: Self-Harness is configured for the most rollouts yet is the weakest, while HarnessFix obtains the strongest baseline results from the smallest baseline budget.
% What orders them is how easily a modification is accepted.
% The baselines score a candidate by its aggregate pass rate on a fixed split, without attributing the change to the modification or checking that untargeted behavior survives.
% Since most tasks in that split do not exercise the targeted behavior, the resulting difference is comparable in size to trial noise, so an edit can be accepted on a chance gain; because accepted edits are cumulative and never revisited, the regressions they carry surface only on TEST.
% More candidates then mean more chances to accept such an edit, which is why the largest configured budget produces the weakest harnesses, and why degradations concentrate on Banking and Terminal-Bench, whose behaviors a shared split covers most unevenly.

% Table~\ref{tab:verification-results} isolates this effect by varying only how the verification batch is selected.

\section{Analysis}
\label{sec:analysis}

% \secref{sec:experiments} shows that \method improves harnesses under a small budget and that the baselines often do not.
% This section asks which part of the method is responsible, what the evolved harnesses actually contain, and how the behavior changes with the budget.
% Unless stated otherwise, analyses use OpenCode as the target harness and the same TRAIN and TEST splits as \secref{sec:experiment-setup}.

To understand how \method achieves effective harness evolution under a limited interaction budget, we analyze two aspects of its behavior-aware verification process. First, we isolate the contributions of behavior-aware batch selection and attributable-evidence gating through ablations. Second, we examine how task-space structure affects the ability to identify modifications that receive consistent, attributable support across related tasks. Appendix~\ref{sec:appendix-decision-analysis} complements these analyses with decision-level evidence on accepted modifications and budget allocation.

\subsection{Ablation: Selection and Gating}
\label{sec:analysis-selection}

\method improves verification through two complementary mechanisms: it selects tasks relevant to each candidate modification and accepts the modification only when the observed improvement can be attributed to it. We ablate these mechanisms independently to isolate their contributions.

\paragraph{Settings.}
All variants use $B=200$ and modify only the verification-stage configuration of \method; all other settings follow the main experiments. To ablate batch selection, \emph{Fixed Batch}, \emph{Random Batch}, and \emph{RHO-based Batch}~\cite{pan2026dark} use a fixed subset, a uniformly resampled subset, and RHO-based task selection, respectively, while retaining the attributable-evidence gate. For these variants, we set the verification batch size to 10, or one third of the 30 TRAIN tasks, yielding a verification fraction comparable to the validation fraction used in HarnessFix~\cite{chen2026harnessfix}. To ablate the acceptance mechanism, \emph{Metric-Only Gate} retains behavior-aware selection but accepts candidates solely according to the aggregate primary metric.

\paragraph{Results.}
Table~\ref{tab:verification-results} shows that removing either mechanism substantially weakens performance. Fixed, random, and RHO-based batches frequently include tasks unrelated to the candidate modification, introducing variation that the diagnostic analysis cannot attribute to the targeted behavior. Consequently, these variants often fail to produce candidates that pass verification. In contrast, \emph{Metric-Only Gate} promotes candidates more readily but also accepts aggregate gains unsupported by attributable evidence; these gains yield no clear held-out improvement on Banking or BIRD. Overall, behavior-aware selection improves the relevance of verification evidence, while attributable-evidence gating prevents noisy or unsupported gains from being retained.

\begin{table}[t] 
\centering \small \setlength{\tabcolsep}{1mm} 

\begin{tabular}{lccc}
\toprule

\vheadl{\textbf{Method}}
& \vheadc{\textbf{$\tau^2$-bench}\\\textbf{Retail}}
& \vheadc{\textbf{$\tau^3$-bench}\\\textbf{Banking}}
& \vheadc{\textbf{BIRD}\\\textbf{Mini-Dev}\\
         \textbf{(Challenging)}} \\

\midrule 

$\mathcal{H}_0$ & 75.00 & 20.90 & 37.50 \\ 
% \midrule 
\cmidrule(lr){1-4} 

Fixed Batch
& 75.00 
& 20.90
& 37.50 \\ 

Random Batch 
& 75.00
& 20.90
& 37.50 \\ 

RHO-based Batch
& 75.00 
& 20.90 
& 38.89 \\ 

Metric-Only Gate 
& 80.00 
& 20.90 
& 37.50 \\ 
% \midrule 
\cmidrule(lr){1-4} 

\textbf{HarnessLens (Ours)} 
& \textbf{85.00} 
& \textbf{25.37} 
& \textbf{45.83} \\ 

\bottomrule 
\end{tabular} 
\caption{Ablation of behavior-aware batch selection and attributable-evidence gating on OpenCode.} 

\label{tab:verification-results} 
\end{table}

\subsection{How Does Task Diversity Affect Evolution?}
\label{sec:analysis-diversity}

The gains on Banking and BIRD suggest that evolution benefits from recurring, actionable weaknesses: related tasks can reinforce a modification across examples. Retail instead starts from a strong harness with less potential for further improvement; its trajectories mainly serve as checks for regressions, so retaining the original harness is often desirable.

\begin{figure}[t]
\centering
\includegraphics[width=0.95\columnwidth]{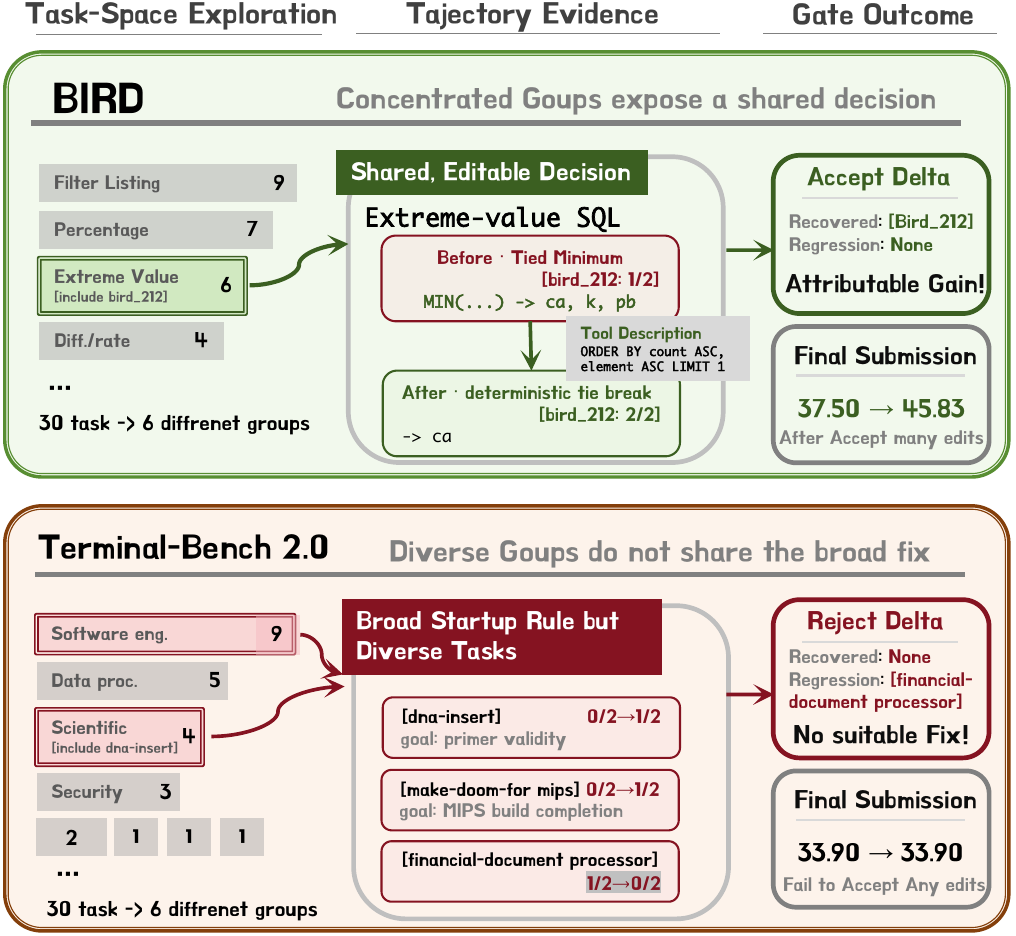}
    \caption{
        Illustrative contrast between task-space structure and evolution
        evidence. In BIRD, related tasks expose a shared extreme-value SQL decision, enabling an attributable recovery and an accepted edit. Terminal-Bench 2.0 contains more distinct task goals and execution paths; the tested broad rule yields no recovery and causes a regression, so it is rejected. The BIRD TEST result is cumulative across all accepted edits.
    }
    \label{fig:task-diversity-evidence}
\end{figure}

Figure~\ref{fig:task-diversity-evidence} contrasts the evidence available in BIRD and Terminal-Bench 2.0. 
In BIRD, several tasks involving extreme-value queries expose the same SQL decision, enabling an attributable recovery.
Terminal tasks involve distinct goals and execution paths; the tested broad rule recovers no tasks and causes an attributable regression, so the gate retains the original harness. 
This is not merely a TRAIN--TEST mismatch: diversity within both splits limits modifications that are supported by TRAIN and broadly applicable to TEST.

This also explains why fixed-set optimizers benefit from repetitive objectives that provide consistent signals toward the same direction. SkillBoost~\cite{lin2026rethinking} similarly finds that optimized skills transfer most naturally between similar tasks. 
Since our objective is held-out generalization rather than exhaustive TRAIN optimization, high task diversity makes broadly effective modifications harder to identify. 
Appendix~\ref{sec:appendix-bird-evolution-example} provides an evolution trace for an OpenCode--BIRD run.

\section{Conclusion}

We introduce \method, a budget-aware framework that evolves agent harnesses through behavior-aware verification. 
By selecting tasks related to each modification and accepting changes based on attributable behavioral evidence, \method uses limited rollouts more effectively while avoiding unsupported
regressions. 
Across three harnesses and four benchmarks, it consistently improves or preserves initial performance and achieves average performance
improvements of 7.6--13.6\%. 
These results show that reliable harness evolution
depends not only on the amount of verification, but also on directing verification toward the behaviors each modification is intended to change.

\section*{Limitations}
Our evaluation covers one model family, three harnesses, and four public benchmarks. While the results consistently support the effectiveness of behavior-aware verification across these settings, its effectiveness under a broader range of models, harness architectures, and open-ended deployment environments has not yet been fully validated. In addition, our budget counts LLM sessions and task trials as auditable interaction units, but does not normalize token usage, latency, or monetary cost across roles and benchmarks.
 % Re-enable before ARR submission.


\bibliography{custom}

@article{lin2026rethinking,
  title   = {Rethinking Self-Evolution: A Constrained
             Exploration-Exploitation Process for Mitigating
             Skill Overfitting},
  author  = {Lin, Hongqiang and Liu, Chao and Bai, Xiaofan and
             Jin, Xuan and Li, Yuhong and Zheng, Nenggan and
             Cao, Xipeng},
  journal = {Computing Research Repository},
  volume  = {arXiv:2607.26643},
  year    = {2026},
  url     = {https://arxiv.org/abs/2607.26643}
}

@article{ning2026code,
  title   = {Code as Agent Harness},
  author  = {Ning, Xuying and
             Tieu, Katherine and
             Fu, Dongqi and
             Wei, Tianxin and
             Li, Zihao and
             Bei, Yuanchen and
             Zou, Jiaru and
             Ai, Mengting and
             Liu, Zhining and
             Li, Ting-Wei and
             Chen, Lingjie and
             Zhao, Yanjun and
             Yang, Ke and
             Li, Bingxuan and
             Qian, Cheng and
             Li, Gaotang and
             Lin, Xiao and
             Zeng, Zhichen and
             Qiu, Ruizhong and
             Chen, Sirui and
             Sun, Yifan and
             Yang, Xiyuan and
             Wang, Ruida and
             Pan, Rui and
             Yang, Chenyuan and
             Zhang, Dylan and
             Fang, Liri and
             Cui, Zikun and
             Cao, Yang and
             Chen, Pan and
             Sun, Dorothy and
             Chen, Ren and
             Srinivasan, Mahesh and
             Mathur, Nipun and
             Xia, Yinglong and
             Li, Hong and
             Yan, Hong and
             Lu, Pan and
             Zhang, Lingming and
             Zhang, Tong and
             Tong, Hanghang and
             He, Jingrui},
  journal = {Computing Research Repository},
  volume  = {arXiv:2605.18747},
  year    = {2026},
  url     = {https://arxiv.org/abs/2605.18747}
}

@article{li2024can,
  title = {Can {LLM}s Already Serve as a Database Interface? A Big Bench for Large-Scale Database Grounded Text-to-{SQL}},
  author = {Li, Jinyang and Hui, Binyuan and Qu, Ge and Yang, Jiaxi and Li, Binhua and Li, Bowen and Wang, Bailin and Qin, Bowen and Geng, Ruiying and Huo, Nan and others},
  journal = {Advances in Neural Information Processing Systems},
  volume = {36},
  year = {2024}
}

@misc{openai2026codex,
  author       = {{OpenAI}},
  title        = {{Codex}: A Lightweight Coding Agent},
  year         = {2026},
  howpublished = {\url{https://github.com/openai/codex}},
  note         = {Accessed: 2026-07-26}
}

@misc{opencode2026,
  author       = {{Anomaly}},
  title        = {{OpenCode}: The Open Source AI Coding Agent},
  year         = {2026},
  howpublished = {\url{https://github.com/anomalyco/opencode}},
  note         = {Accessed: 2026-07-26}
}

@misc{zechner2025pi,
  author       = {Zechner, Mario},
  title        = {{Pi} Agent Harness},
  year         = {2025},
  howpublished = {\url{https://github.com/earendil-works/pi}},
  note         = {Accessed: 2026-07-26}
}

@article{huang2026memoharness,
  title   = {{MemoHarness}: Agent Harnesses That Learn from Experience},
  author  = {Yue Huang and Wenjie Wang and Han Bao and Yuchen Ma
             and Xiaonan Luo and Yi Nian and Haomin Zhuang
             and Zheyuan Liu and Yue Zhao and Xiangliang Zhang},
  journal = {Computing Research Repository},
  volume  = {arXiv:2607.14159},
  year    = {2026},
  url     = {https://arxiv.org/abs/2607.14159}
}

@article{lee2026rhi,
  title   = {Recursive Harness Self-Improvement},
  author  = {Hyunin Lee and Jinglue Xu and Jeffrey Seely
             and Donghyun Lee and Matei Zaharia and Yujin Tang},
  journal = {Computing Research Repository},
  volume  = {arXiv:2607.15524},
  year    = {2026},
  url     = {https://arxiv.org/abs/2607.15524}
}

@article{du2026livingharness,
  title   = {{Living-Harness} Is an Interactive-Agent Evolver},
  author  = {Yuetian Du and Yucheng Wang and He Xu and Jiexu Xu
             and Shanwen Tan and Bing Zhao and Boyu Yang and Zhijie Xu
             and Ming Kong and Hu Wei and Jie Liu and Qiang Zhu},
  journal = {Computing Research Repository},
  volume  = {arXiv:2607.26598},
  year    = {2026},
  url     = {https://arxiv.org/abs/2607.26598}
}

@article{liu2026adaptiveharness,
  title   = {Adaptive Auto-Harness: Sustained Self-Improvement for
             Agentic System Deployment on Open-Ended Task Streams},
  author  = {Zewen Liu and Zhan Shi and Yisi Sang and Bing He
             and Minhua Lin and Tianxin Wei and Dakuo Wang
             and Benoit Dumoulin and Wei Jin and Hanqing Lu},
  journal = {Computing Research Repository},
  volume  = {arXiv:2606.01770},
  year    = {2026},
  url     = {https://arxiv.org/abs/2606.01770}
}

@article{karten2026continualharness,
  title   = {Continual Harness: Online Adaptation for
             Self-Improving Foundation Agents},
  author  = {Seth Karten and Joel Zhang and Tersoo Upaa Jr.
             and Ruirong Feng and Wenzhe Li and Chengshuai Shi
             and Chi Jin and Kiran Vodrahalli},
  journal = {Computing Research Repository},
  volume  = {arXiv:2605.09998},
  year    = {2026},
  url     = {https://arxiv.org/abs/2605.09998}
}

@inproceedings{agrawal2026gepa,
  title     = {{GEPA}: Reflective Prompt Evolution Can Outperform
               Reinforcement Learning},
  author    = {Lakshya A. Agrawal and Shangyin Tan and Dilara Soylu
               and Noah Ziems and Rishi Khare and Krista Opsahl-Ong
               and Arnav Singhvi and Herumb Shandilya and Michael J. Ryan
               and Meng Jiang and Christopher Potts and Koushik Sen
               and Alexandros G. Dimakis and Ion Stoica and Dan Klein
               and Matei Zaharia and Omar Khattab},
  booktitle = {The Fourteenth International Conference on Learning
               Representations},
  year      = {2026},
  eprint    = {2507.19457},
  archivePrefix = {arXiv},
  primaryClass  = {cs.LG}
}

@inproceedings{zhang2025aflow,
  title     = {{AFlow}: Automating Agentic Workflow Generation},
  author    = {Jiayi Zhang and Jinyu Xiang and Zhaoyang Yu
               and Fengwei Teng and Xiong-Hui Chen and Jiaqi Chen
               and Mingchen Zhuge and Xin Cheng and Sirui Hong
               and Jinlin Wang and Bingnan Zheng and Bang Liu
               and Yuyu Luo and Chenglin Wu},
  booktitle = {The Thirteenth International Conference on Learning
               Representations},
  year      = {2025},
  url       = {https://openreview.net/forum?id=z5uVAKwmjf}
}

@article{chen2026harnessfix,
  title   = {From Failed Trajectories to Reliable {LLM} Agents:
             Diagnosing and Repairing Harness Flaws},
  author  = {Mengzhuo Chen and Junjie Wang and Zhe Liu
             and Yawen Wang and Qing Wang},
  journal = {Computing Research Repository},
  volume  = {arXiv:2606.06324},
  year    = {2026},
  url     = {https://arxiv.org/abs/2606.06324}
}

@article{pan2026dark,
  title   = {Evolving Agents in the Dark: Retrospective Harness
             Optimization via Self-Preference},
  author  = {Wenbo Pan and Shujie Liu and Chin-Yew Lin
             and Jingying Zeng and Xianfeng Tang and Xiangyang Zhou
             and Yan Lu and Xiaohua Jia},
  journal = {Computing Research Repository},
  volume  = {arXiv:2606.05922},
  year    = {2026},
  url     = {https://arxiv.org/abs/2606.05922}
}

@article{shi2026tau,
  title   = {$\tau$-Knowledge: Evaluating Conversational Agents over
             Unstructured Knowledge},
  author  = {Shi, Quan and Zytek, Alexandra and Razavi, Pedram
             and Narasimhan, Karthik and Barres, Victor},
  journal = {Computing Research Repository},
  volume  = {arXiv:2603.04370},
  year    = {2026},
  url     = {https://arxiv.org/abs/2603.04370}
}

@article{barres2025tau2,
  title         = {$\tau^2$-Bench: Evaluating Conversational Agents in a
                   Dual-Control Environment},
  author        = {Barres, Victor and Dong, Honghua and Ray, Soham
                   and Si, Xujie and Narasimhan, Karthik},
  journal       = {Computing Research Repository},
  volume        = {arXiv:2506.07982},
  year          = {2025},
  url           = {https://arxiv.org/abs/2506.07982}
}

@article{luo2026gsme,
  title   = {Self-Evolving Agent Harnesses via Gated Semantic
             Quality-Diversity},
  author  = {Luo, Xiaotian and Wang, Fengxingyu and Hu, Chuanrui
             and Xue, Dizhan and Deng, Yafeng},
  journal = {Computing Research Repository},
  volume  = {arXiv:2607.13683},
  year    = {2026},
  url     = {https://arxiv.org/abs/2607.13683}
}

@article{hao2026recreate,
  title = {{ReCreate}: Reasoning and Creating Domain Agents Driven by Experience},
  author = {Hao, Zhezheng and Wang, Hong and Luo, Jian and Zhang, Jianqing and Zhou, Yuyan and Lin, Qiang and Wang, Can and Dong, Hande},
  journal = {Computing Research Repository},
  volume = {arXiv:2601.11100},
  year = {2026},
  url = {https://arxiv.org/abs/2601.11100}
}

@article{lou2026autoharness,
  title = {{AutoHarness}: improving {LLM} agents by automatically synthesizing a code harness},
  author = {Lou, Xinghua and L{\'a}zaro-Gredilla, Miguel and Dedieu, Antoine and Wendelken, Carter and Lehrach, Wolfgang and Murphy, Kevin P.},
  journal = {Computing Research Repository},
  volume = {arXiv:2603.03329},
  year = {2026},
  url = {https://arxiv.org/abs/2603.03329}
}

@article{lee2026metaharness,
  title = {{Meta-Harness}: End-to-End Optimization of Model Harnesses},
  author = {Lee, Yoonho and Nair, Roshen and Zhang, Qizheng and Lee, Kangwook and Khattab, Omar and Finn, Chelsea},
  journal = {Computing Research Repository},
  volume = {arXiv:2603.28052},
  year = {2026},
  url = {https://arxiv.org/abs/2603.28052}
}

@article{lin2026ahe,
  title = {Agentic Harness Engineering: Observability-Driven Automatic Evolution of Coding-Agent Harnesses},
  author = {Lin, Jiahang and Liu, Shichun and Pan, Chengjun and Lin, Lizhi and Dou, Shihan and Xi, Zhiheng and Huang, Xuanjing and Yan, Hang},
  journal = {Computing Research Repository},
  volume = {arXiv:2604.25850},
  year = {2026},
  url = {https://arxiv.org/abs/2604.25850}
}

@article{cai2026moss,
  title = {{MOSS}: Self-Evolution through Source-Level Rewriting in Autonomous Agent Systems},
  author = {Cai, Qianshu and Zhang, Yonggang and Jia, Xianzhang and Zheng, Huajiang and Xue, Wei and Song, Jun and Tian, Xinmei and Guo, Yike},
  journal = {Computing Research Repository},
  volume = {arXiv:2605.22794},
  year = {2026},
  url = {https://arxiv.org/abs/2605.22794}
}

@article{zhang2026selfharness,
  title = {Self-Harness: Harnesses That Improve Themselves},
  author = {Zhang, Hangfan and Zhang, Shao and Li, Kangcong and Zhang, Chen and Chen, Yang and Zhang, Yiqun and Bai, Lei and Hu, Shuyue},
  journal = {Computing Research Repository},
  volume = {arXiv:2606.09498},
  year = {2026},
  url = {https://arxiv.org/abs/2606.09498}
}

@article{nie2026tthe,
  title = {{TTHE}: Test-Time Harness Evolution},
  author = {Nie, Jun and Zhang, Yonggang and Song, Jun and Cai, Qianshu and Yu, Dahai and Guo, Yike and Tian, Xinmei and Han, Bo},
  journal = {Computing Research Repository},
  volume = {arXiv:2607.08124},
  year = {2026},
  url = {https://arxiv.org/abs/2607.08124}
}

@inproceedings{yang2024opro,
  title = {Large Language Models as Optimizers},
  author = {Yang, Chengrun and Wang, Xuezhi and Lu, Yifeng and Liu, Hanxiao and Le, Quoc V. and Zhou, Denny and Chen, Xinyun},
  booktitle = {International Conference on Learning Representations},
  year = {2024},
  eprint = {2309.03409},
  archivePrefix = {arXiv},
  primaryClass = {cs.LG}
}

@inproceedings{opsahlong2024mipro,
  title = {Optimizing Instructions and Demonstrations for Multi-Stage Language Model Programs},
  author = {Opsahl-Ong, Krista and Ryan, Michael J. and Purtell, Josh and Broman, David and Potts, Christopher and Zaharia, Matei and Khattab, Omar},
  booktitle = {Proceedings of the 2024 Conference on Empirical Methods in Natural Language Processing},
  year = {2024},
  eprint = {2406.11695},
  archivePrefix = {arXiv},
  primaryClass = {cs.CL}
}

@article{yuksekgonul2024textgrad,
  title = {{TextGrad}: Automatic ``Differentiation'' via Text},
  author = {Yuksekgonul, Mert and Bianchi, Federico and Boen, Joseph and Liu, Sheng and Huang, Zhi and Guestrin, Carlos and Zou, James},
  journal = {Computing Research Repository},
  volume = {arXiv:2406.07496},
  year = {2024},
  url = {https://arxiv.org/abs/2406.07496}
}

@article{merrill2026terminalbench,
  title = {Terminal-Bench: Benchmarking Agents on Hard, Realistic Tasks in Command Line Interfaces},
  author = {Merrill, Mike A. and Shaw, Alexander G. and Carlini, Nicholas and Li, Boxuan and Raj, Harsh and others},
  journal = {Computing Research Repository},
  volume = {arXiv:2601.11868},
  year = {2026},
  url = {https://arxiv.org/abs/2601.11868}
}

@article{deepseek2026v4,
  title = {{DeepSeek-V4}: Towards Highly Efficient Million-Token Context Intelligence},
  author = {DeepSeek-AI and others},
  journal = {Computing Research Repository},
  volume = {arXiv:2606.19348},
  year = {2026},
  url = {https://arxiv.org/abs/2606.19348}
}

\clearpage
\appendix

\section{Implementation and Verification Details}
\label{sec:appendix-reproducibility}

This appendix provides implementation details for the three components of \method described in \secref{sec:method}: Context Exploration, Trajectory Diagnosis, and Harness Evolution. 
It specifies the controller procedure, budget accounting, role interfaces, editable harness components, and verification protocol used in our experiments.

\subsection{Controller Procedure and Budget Accounting}
\label{sec:appendix-procedure}
\label{sec:appendix-budget}

Algorithm~\ref{alg:harnesslens-operational} instantiates the procedure in \secref{sec:method}. 
Each candidate is constructed from the current confirmed
harness, so accepted modifications accumulate and rejected candidates leave the current harness unchanged. 
All model-based roles run in fresh sessions and communicate only through schema-validated artifacts.

\begin{algorithm}[t]
\small
\caption{Controller-level implementation of \method.}
\label{alg:harnesslens-operational}
\begin{algorithmic}[1]
\REQUIRE $\mathcal{H}_0$, $\mathcal{T}_{\mathrm{train}}$, $B$, $K=2$
\ENSURE Confirmed harness $\mathcal{H}$
\STATE Run the initial rollout of $\mathcal{H}_0$ on
$\mathcal{T}_{\mathrm{train}}$ with $K$ trials per task
\STATE Run Task-Space and Harness-Space Exploration
\STATE Apply Trajectory Diagnosis to the initial trajectories to obtain proposals
\STATE $\mathcal{H}\gets\mathcal{H}_0$
\WHILE{a supported proposal and a complete verification cycle remain}
  \STATE Evolution agent selects proposal $p$ and batch $S$, $|S|\geq5$
  \STATE $\widetilde{\mathcal{H}} \gets \operatorname{Edit}(\mathcal{H},p)$
  \IF{the modified component passes the runtime check}
    \IF{$\mathcal{H}=\mathcal{H}_0$}
      \STATE Reuse the initial trajectories for $\mathcal{H}_0$ on $S$
    \ELSE
      \STATE Evaluate $\mathcal{H}$ on $S$
    \ENDIF
    \STATE Evaluate $\widetilde{\mathcal{H}}$ on $S$ with matched trials
    \STATE Apply Trajectory Diagnosis to the paired trajectories
    \STATE Evolution agent reviews the verification evidence and metrics
    \IF{attributable positive evidence and no attributable regression}
      \STATE Construct confirmation batch $S'$ (\secref{sec:appendix-task-selection})
      \STATE Evaluate both harnesses on $S'$ under fresh matched conditions
      \STATE Apply Trajectory Diagnosis to the confirmation trajectories
      \STATE Evolution agent reviews the confirmation evidence and metrics
      \IF{supported evidence remains and the primary metric improves}
        \STATE $\mathcal{H}\gets\widetilde{\mathcal{H}}$
      \ENDIF
    \ENDIF
  \ENDIF
  \STATE Add revised or newly diagnosed proposals
  \STATE Update the evidence history and remaining budget
\ENDWHILE
\RETURN $\mathcal{H}$
\end{algorithmic}
\end{algorithm}

Every task rollout and LLM session consumes one unit of the interaction budget $B$, as specified in \secref{sec:optimization-objective}. 
With 30 TRAIN tasks and $K=2$ trials per task, the initial rollout costs $30K=60$ units. 
For a verification batch of $n$ tasks against the current harness $\mathcal{H}_j$, the required cost is
\[
\begin{aligned}
C_{1}(n,\mathcal{H}_j)
&=6+nK\bigl(1+\mathbb{I}[\mathcal{H}_j\neq\mathcal{H}_0]\bigr)\\
&\quad+2\left\lceil\frac{n}{3}\right\rceil ,
\end{aligned}
\]
The fixed six units cover proposal selection, candidate editing, the runtime check, two analyses of intended effects and regressions, and the final review. 
The rollout term covers candidate trials and, when $\mathcal{H}_j\neq\mathcal{H}_0$, fresh trials under the current harness. 
The final term covers independently reviewed behavior comparisons in bundles of at most three tasks.
The confirmation round always collects fresh trials under both harnesses and costs
\[
C_{2}(n)
=3+2nK+2\left\lceil\frac{n}{3}\right\rceil .
\]
Before an iteration starts, the controller checks that the remaining budget covers both rounds and a three-unit retry buffer. 
Invalid structured outputs may be retried once for exploration and twice for diagnosis or review; every retry remains charged. 
If the remaining budget cannot support both rounds, the controller does not start another candidate iteration.

For example, with $K=2$, a five-task verification against
$\mathcal{H}_0$ costs 20 units because its initial trajectories are reused, and the corresponding confirmation round costs 27 units. 
If the current harness differs from $\mathcal{H}_0$, a six-task verification and confirmation round cost 34 and 31 units, respectively. 
Including the three-unit retry
buffer, this six-task iteration requires $34+31+3=68$ available units. 
The controller starts an iteration only when the complete two-round cycle is affordable and returns the current confirmed harness otherwise.

\subsection{Role Interfaces}
\label{sec:appendix-role-boundaries}

Table~\ref{tab:role-boundaries} summarizes the visible inputs, excluded information, and outputs of the model-based roles used by \method. 
Within Trajectory Diagnosis, behavior comparison is performed without access to the candidate diff; the subsequent assessment of intended effects and regressions
receives the diff only after that comparison is complete.

\begin{table*}[t]
\centering
\small
\setlength{\tabcolsep}{5pt}
\renewcommand{\arraystretch}{1.12}
\begin{tabular}{@{}p{0.18\textwidth}p{0.30\textwidth}p{0.24\textwidth}p{0.20\textwidth}@{}}
\toprule
\textbf{Role} & \textbf{Visible information} & \textbf{Excluded information} & \textbf{Output} \\
\midrule
Task-Space Exploration & TRAIN queries, public policies, and tool schemas & Trajectories, rewards, candidates, and TEST & Task groups \\
\addlinespace[3pt]
Harness-Space Exploration & Harness documentation and runtime observations & Task outcomes and domain-specific proposals & Editable-component inventory \\
\addlinespace[3pt]
Experience Extraction & Model-visible TRAIN trajectories and outcomes & Candidate diff and TEST & Reusable experiences and recurring deficiencies \\
\addlinespace[3pt]
Experience Analysis & Extracted evidence, task groups, and component inventory & TEST and hidden references & Evidence-supported modification proposals \\
\addlinespace[3pt]
Behavior Comparison & Current-harness and candidate trajectories & Candidate diff and hidden evaluator explanations & Task-level behavior changes \\
\addlinespace[3pt]
Trajectory Diagnosis (verification) & Comparison reports, task groups, and candidate diff & TEST and hidden references & Assessment of intended effects and regressions \\
\addlinespace[3pt]
Harness Editor & One proposal, the component inventory, and an isolated snapshot of the current harness & TEST, credentials, evaluator state, and fixed runtime limits & Candidate snapshot and file-level diff \\
\addlinespace[3pt]
Evolution Agent & Proposals, evidence history, task-selection records, and budget state & TEST and hidden references & Proposal selection and candidate acceptance, rejection, or revision \\
\bottomrule
\end{tabular}
\caption{Information boundaries between the model-based implementation roles.
Budget arithmetic and trial counts are controller-owned rather than accepted
from model outputs.}
\label{tab:role-boundaries}
\end{table*}

Task- and Harness-Space Exploration use at most 30 agent steps, the Harness Editor at most 40, and the Trajectory Diagnosis and evolution-agent roles at most 60. 
Their complete prompts and output schemas are included in the
anonymized software supplement.
All roles use the same high-reasoning profile as the evaluated agents, with a 65,536-token context limit and a 24,576-token output limit. 
Exploration sessions have a 1,800-second wall-clock limit and the remaining roles have a 3,600-second limit. 

\subsection{Harness Search Spaces}
\label{sec:appendix-harness-spaces}

Each candidate is an isolated snapshot of the current configurable state \(\mathbf{h}_j\) and the task project. The harness framework, base model, provider, permissions, benchmark tools, and evaluator state remain fixed. 
Patches to tool and parameter descriptions must bind to exact names already exposed by the environment and therefore cannot introduce new tools.

\begin{table*}[t]
\centering
\small
\setlength{\tabcolsep}{5pt}
\renewcommand{\arraystretch}{1.12}
\begin{tabular}{@{}p{0.10\textwidth}p{0.36\textwidth}p{0.22\textwidth}p{0.24\textwidth}@{}}
\toprule
\textbf{Harness} & \textbf{User-configurable components} & \textbf{Visibility} & \textbf{Main regression risks} \\
\midrule
OpenCode & Instructions, skills, tool and parameter descriptions, agent definitions, commands, reference sources, and compaction configuration & Startup, tool schema, skill loading, or lifecycle events & Broad prompt effects, context growth, inert files, or authority changes \\
\addlinespace[3pt]
Codex & Developer and project instructions, skills, lifecycle-hook context, tool and parameter descriptions, and compaction prompt & Startup, tool schema, skill loading, or lifecycle events & Broad instructions, trigger failure, repeated injection, or context loss \\
\addlinespace[3pt]
Pi & System-prompt append, project instructions, skills, tool and parameter descriptions, and compaction configuration & Startup, tool schema, on-demand reading, or lifecycle events & Native-prompt replacement, trust dependence, trigger failure, or context loss \\
\bottomrule
\end{tabular}
\caption{User-configurable components identified by Harness-Space Exploration
for the three evaluated harnesses.}
\label{tab:harness-components}
\end{table*}

Before verification, a one-trial load check records the effective startup instructions, discovered skill metadata, tool schema, and lifecycle configuration. 
A candidate is rejected without rollout if a modified component has no applicable checker or its content is absent from the effective runtime context. 
This prevents edits to inert files from being credited as behavioral changes.

\subsection{Verification and Update Protocol}
\label{sec:appendix-behavior-verification}

\subsubsection{Task Selection}
\label{sec:appendix-task-selection}

The evolution agent assigns each task in an initial verification batch one of the four selection labels in Table~\ref{tab:verification-task-roles}. 
The labels record each task's role in testing the targeted behavior and detecting potential regressions.

\begin{table}[t]
\centering
\small
\setlength{\tabcolsep}{5pt}
\renewcommand{\arraystretch}{1.12}
\begin{tabular}{@{}lp{0.64\columnwidth}@{}}
\toprule
\textbf{Selection label} & \textbf{Criterion} \\
\midrule
Conversion & Initial failure that directly exercises the proposed behavior \\
\addlinespace[2pt]
Positive control & Initial success directly linked to supporting evidence \\
\addlinespace[2pt]
Preservation & Other initial success used to probe collateral regression \\
\addlinespace[2pt]
Diagnostic & Other initial failure used to test the proposal's scope \\
\bottomrule
\end{tabular}
\caption{Controller records used to construct a verification batch.}
\label{tab:verification-task-roles}
\end{table}

A verification batch contains at least five distinct TRAIN tasks, including at least one conversion task and one task linked to the proposal's supporting trajectory. 
Remaining positions cover related task groups, affected tools, and preservation risks. 
The evolution agent selects the tasks and provides their
selection labels and evidence links; the controller validates task distinctness, batch size, trial count, and affordability.

The confirmation round has the same batch size but uses mostly fresh tasks. 
It retains at most $\min(2,n-1)$ verification tasks that require further confirmation and fills the remaining positions with previously unused TRAIN tasks that succeeded under the initial harness. 
The controller prioritizes task groups not represented in the verification and reuses other verification tasks only when fresh eligible tasks are exhausted.

\subsubsection{Paired Rollouts and Review}
\label{sec:appendix-promotion}

Each round uses two paired trials per task. 
Trials under the current and candidate harnesses share benchmark-controlled conditions within a round, while
the confirmation round uses fresh trial seeds. 
For interactive benchmarks, the user-simulator seed is determined by the domain, task ID, and trial index;
target-agent seeds additionally include an agent namespace. 
Pairing reduces avoidable variation but does not assume provider-side determinism. 

Based on the binary rewards defined in \secref{sec:optimization-objective}, behavior comparison assigns each task one of the labels in Table~\ref{tab:comparison-status}.

\begin{table}[t]
\centering
\small
\setlength{\tabcolsep}{5pt}
\renewcommand{\arraystretch}{1.12}
\begin{tabular}{@{}lp{0.65\columnwidth}@{}}
\toprule
\textbf{Status} & \textbf{Definition} \\
\midrule
Recovered & The current harness has no successful trial and the candidate has at least one \\
\addlinespace[2pt]
Stable success & The current harness has a successful trial and all candidate trials succeed \\
\addlinespace[2pt]
Regressed & The current harness has a successful trial and the candidate has none \\
\addlinespace[2pt]
Still failing & Neither harness has a successful trial \\
\addlinespace[2pt]
Mixed & All remaining split or partial outcomes \\
\bottomrule
\end{tabular}
\caption{Task-level labels used by behavior comparison.}
\label{tab:comparison-status}
\end{table}

Comparison reports are independently checked against the source trajectories before verification-stage Trajectory Diagnosis receives the candidate diff. 
A verification stage advances only if it contains attributable positive evidence and no attributable regression. 
Positive evidence is either an attributed recovery or an attributed stable success with an increased pass count; preservation alone is insufficient. 
An update additionally requires improvement under the primary metric on the confirmation batch and a final accept decision. 
Evidence from an incomplete or unaffordable confirmation round can inform later proposals but cannot update the confirmed harness.

\section{Experimental Protocol}
\label{sec:appendix-experimental-config}

\subsection{Data Splits}

Table~\ref{tab:appendix-splits} gives the fixed split sizes and sampling
provenance. The exact ordered task IDs, sampling seed (42), and upstream
benchmark revisions are included in the anonymized software supplement.

\begin{table*}[t]
\centering
\small
\setlength{\tabcolsep}{6pt}
\renewcommand{\arraystretch}{1.15}
\begin{tabular}{@{}lrrp{0.50\textwidth}@{}}
\toprule
\textbf{Benchmark} & \textbf{TRAIN} & \textbf{TEST} & \textbf{Sampling record} \\
\midrule
$\tau^2$-bench Retail & 30 & 40 & Official 40-task TEST split; TRAIN drawn with seed 42 from the other 74 tasks \\
\addlinespace[3pt]
$\tau^3$-bench Banking Knowledge & 30 & 67 & TRAIN drawn with seed 42 from all 97 tasks; TEST is the remainder \\
\addlinespace[3pt]
Terminal-Bench 2.0 & 30 & 59 & TRAIN drawn with seed 42 from all 89 tasks; TEST is the remainder \\
\addlinespace[3pt]
BIRD Mini-Dev (Challenging) & 30 & 72 & TRAIN drawn with seed 42 from the 102 challenging items in official JSONL order; TEST is the remainder \\
\bottomrule
\end{tabular}
\caption{Fixed splits used for harness evolution and blind evaluation. Exact
ordered IDs and upstream revisions are distributed in the anonymized software
supplement.}
\label{tab:appendix-splits}
\end{table*}

\subsection{Models, Harnesses, and Runtime Limits}

All evaluated agents and model-based roles use \texttt{deepseek-v4-flash-preview}. 
The evaluated harness versions are OpenCode 1.17.13, Codex CLI 0.144.4, and Pi Coding Agent 0.80.10. 
Banking Knowledge uses BM25 retrieval.

For $\tau^2$ interaction, the agent may make at most 10 tool calls per conversation turn, 40 conversation turns, and 60 tool calls overall; the per-turn timeout is 180 seconds. 
The user simulator uses temperature 0.3 and is reset with the stable pairing seed for each trial. 
Terminal-Bench uses at most 50 agent steps, a 600-second agent timeout, and a 1,800-second verifier timeout. 
BIRD uses at most 30 tool calls per round, a 600-second agent timeout, a 5-second query timeout, and a 30-second grader timeout.

Evaluator-owned overrides are applied after candidate-native configuration is loaded. 
They fix the model, provider, task tools, inference limits, and permissions, so a modification cannot change the model, enlarge its own step or token budget, enable external information retrieval, or alter task tools.

\subsection{Metrics and Blind TEST Protocol}

TRAIN uses $K=2$ trials per task. Its controller metric is the trial-level pass rate
\begin{equation}
\operatorname{PassRate}_{\mathrm{TRAIN}}
=\frac{1}{NK}\sum_{i=1}^{N}\sum_{r=1}^{K}R_{i,r},
\end{equation}
which is used only for update decisions alongside the attributable-evidence gate. 
Table~\ref{tab:main-results} instead reports held-out TEST pass@1: one fresh trial per task under the final selected harness. 
We use the term pass@1 only for this single-trial TEST metric.

Initial and evolved harnesses are evaluated on the same ordered TEST split, with the same runtime limits and pairing offsets. 
TEST runs through a separate entry point after evolution completes. 
All information and services associated with TEST---including task IDs, trajectories, evaluation feedback, and rollout services---are unavailable to every exploration, diagnosis, proposal, verification, and review role.

\subsection{Baseline Adaptation and Budget Comparability}
\label{sec:appendix-baselines}

All methods use the same initial harness, 30-task TRAIN split, benchmark runtime, verifier, and mechanisms for applying changes to user-configurable components.
Adaptations are restricted to environment interfaces and do not change a baseline's proposal or selection policy. 
Table~\ref{tab:baseline-protocols} gives the configured protocols; their actual consumption can be lower because of early stopping, retries, or method-specific validation.

\begin{table*}[t]
\centering
\small
\setlength{\tabcolsep}{5pt}
\renewcommand{\arraystretch}{1.15}
\begin{tabular}{@{}lccccp{0.25\textwidth}@{}}
\toprule
\textbf{Method} & \textbf{Internal split} & \textbf{Trials/eval.} & \textbf{Max. iterations} & \textbf{Candidates/iter.} & \textbf{Additional rule} \\
\midrule
Self-Harness & 20/10 & 2 & 20 & 3 & Released held-in/held-out protocol \\
\addlinespace[3pt]
Meta-Harness & 30/0 & 2 & 10 & 1 & All TRAIN tasks used for validation \\
\addlinespace[3pt]
HarnessFix & 20/10 & 1 & 3 & 1 & At most two modification retries \\
\addlinespace[3pt]
\method & 30 (targeted) & 2 & -- & -- & Total budget $B=200$ units \\
\bottomrule
\end{tabular}
\caption{Configured evolution protocols. The internal split is carved from
the same 30 TRAIN tasks; TEST is excluded from every evolution budget.}
\label{tab:baseline-protocols}
\end{table*}

For Self-Harness, 20 iterations each evaluate the current harness and three
candidates on all 30 tasks with two trials, giving
$20\times4\times30\times2=4{,}800$ TRAIN rollouts. Meta-Harness evaluates
the initial harness once and one candidate in each of ten iterations, giving
$11\times30\times2=660$ rollouts. HarnessFix performs three iterations; each
evaluates the current harness and candidate on 30 tasks and rechecks at most two repairs on
20 tasks, giving $3\times(2\times30+2\times20)=300$ rollouts.

\begin{table*}[t]
\centering
\small
\setlength{\tabcolsep}{6pt}
\renewcommand{\arraystretch}{1.15}
\begin{tabular}{@{}lp{0.62\textwidth}r@{}}
\toprule
\textbf{Method} & \textbf{Configured calculation} & \textbf{Maximum} \\
\midrule
Self-Harness & 20 iterations $\times$ 4 evaluations $\times$ 30 tasks $\times$ 2 trials & 4,800 \\
\addlinespace[3pt]
Meta-Harness & 11 evaluations $\times$ 30 tasks $\times$ 2 trials & 660 \\
\addlinespace[3pt]
HarnessFix & 3 iterations $\times$ [2 full evaluations $\times$ 30 tasks $+$ 2 repair rechecks $\times$ 20 tasks] & 300 \\
\addlinespace[3pt]
\method & $B=200$ total units, including 60 initial trials and all LLM sessions & $\leq 200$ \\
\bottomrule
\end{tabular}
\caption{Configured TRAIN cost.
\method uses strictly fewer than 200 task rollouts. The displayed maxima
compare auditable interaction budgets, not matched total compute costs.}
\label{tab:budget}
\end{table*}

\subsection{Artifact Release and Responsible Reporting}
\label{sec:appendix-responsible-reporting}
The anonymized supplement includes our implementation, configurations, prompts, split identifiers, and derived run metadata. 
Third-party agent source code, benchmark data, model weights, and full trajectories are not redistributed and should be obtained from their original releases.

The benchmark dependencies are used under their respective licenses: MIT for $\tau$-bench, Apache License 2.0 for Terminal-Bench, and CC BY-SA 4.0 for BIRD Mini-Dev. 
The agent software and model APIs remain subject to their respective licenses and terms of use.

\paragraph{AI Assistance Disclosure.}
The authors used general-purpose AI assistants for literature search, implementation support, and manuscript drafting and revision. 
The authors reviewed, edited, and verified all resulting code, citations, analyses, and prose, and remain responsible for the final paper.

\section{Supporting Analyses}

\subsection{OpenCode--BIRD Evolution Trace}
\label{sec:appendix-bird-evolution-example}

\begin{figure*}[t]
\centering
\includegraphics[width=\textwidth]{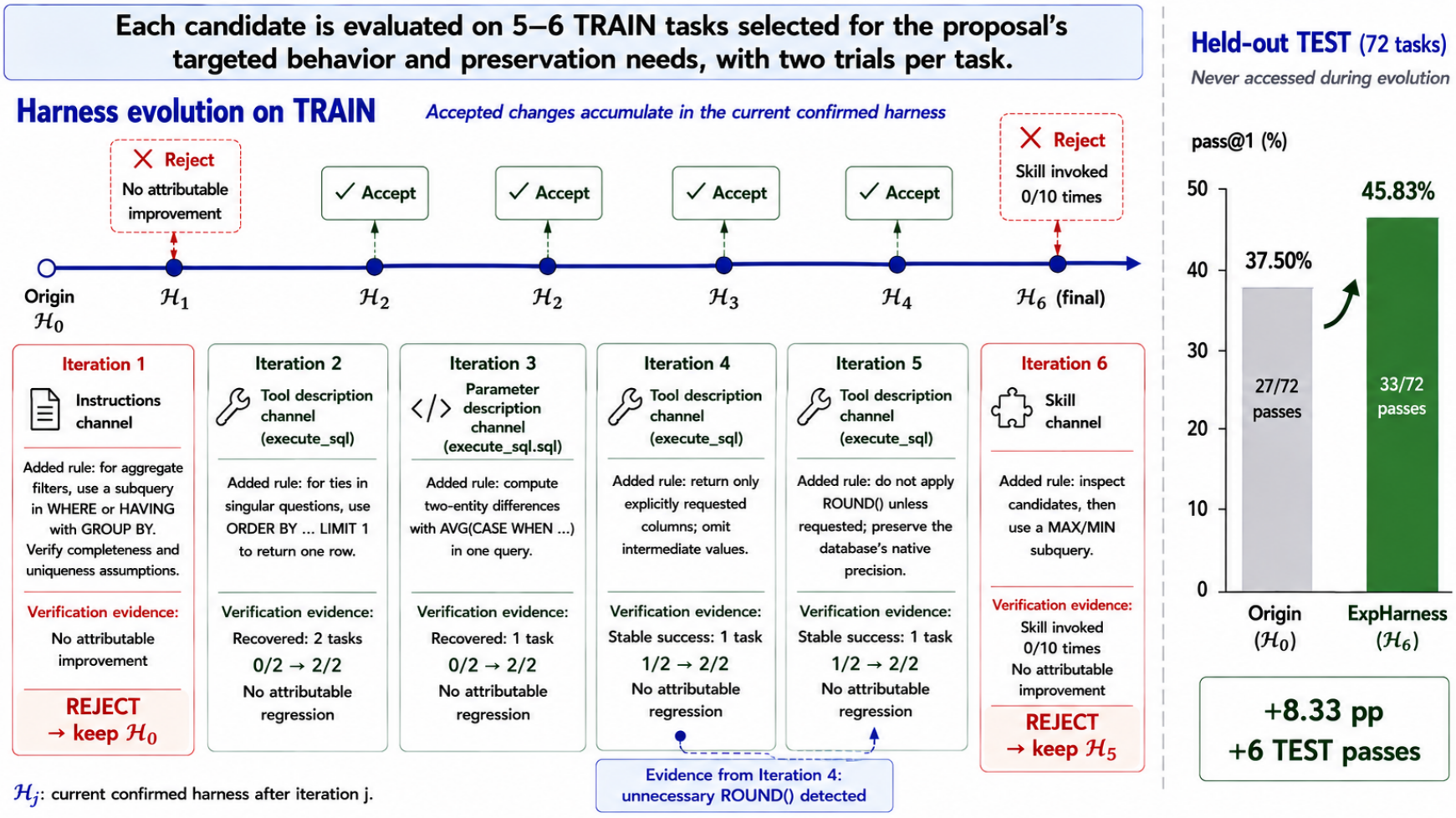}
\caption{Illustrative OpenCode--BIRD evolution trace. 
Four modifications with attributable improvements are accumulated, while two modifications without attributable evidence are rejected. 
The final harness improves held-out TEST pass@1 by 8.33 percentage points.}
\label{fig:appendix-bird-evolution-example}
\end{figure*}

Figure~\ref{fig:appendix-bird-evolution-example} shows the accepted and rejected modifications from an illustrative OpenCode--BIRD run. 
The first instruction edit produced no attributable improvement and was rejected. 
Four later edits added deterministic tie handling, a conditional-aggregation pattern for two-entity differences, an output-column restriction, and native precision guidance. 
For each edit, both verification rounds supported the targeted improvement without an attributable regression, and the primary metric improved on the confirmation batch; the edit was therefore accumulated into the confirmed harness. 
The final skill candidate was not invoked during verification and produced no attributable change, so it was rejected.
Because successive candidates use different modification-specific batches, their local verification rates should not be interpreted as a performance curve. 
The final harness improves held-out TEST pass@1 from 27/72 (37.50\%) to 33/72 (45.83\%), an 8.33 percentage-point increase.

\subsection{Decision-Level Analysis of the OpenCode Runs}
\label{sec:appendix-decision-analysis}

\subsubsection{Scope}
\label{sec:appendix-analysis-pool}

This appendix examines the decisions within the four evolution runs that produce the OpenCode results in Table~\ref{tab:main-results}: one run with $B=200$ for each benchmark. 
Fixing the target harness to OpenCode follows the setting of \secref{sec:analysis-selection} and avoids differences in component inventory across harnesses. 

Table~\ref{tab:appendix-analysis-pool} lists the four runs. 
Three accepted at least one modification, and one returned \(\mathcal{H}_0\). 
Together they contain 21 candidate iterations, 19 of which completed a paired verification comparison against the current confirmed harness. Evidence about TEST outcomes comes from Tables~\ref{tab:main-results} and~\ref{tab:verification-results}, not from this analysis.

\begin{table}[t]
\centering
\small
\setlength{\tabcolsep}{1.3mm}
\begin{tabular}{lcccc}
\toprule
\textbf{Benchmark} & \textbf{Iter.} & \textbf{Units}
& \(\mathcal{H}_0\) & \textbf{Final} \\
\midrule
Retail   & 6 & 196 & 75.00 & 85.00 \\
Banking  & 6 & 197 & 20.90 & 25.37 \\
BIRD     & 5 & 197 & 37.50 & 45.83 \\
Terminal & 4 & 181 & 33.90 & 33.90 \\
\bottomrule
\end{tabular}
\caption{Summary of the four OpenCode evolution runs, one per benchmark.
\textbf{Iter.} reports the number of candidate iterations, and \textbf{Units} reports the interaction-budget units consumed from the total budget of 200.
\(\mathcal{H}_0\) and \textbf{Final} report held-out TEST pass@1 (\%) before and after evolution, respectively. 
The Terminal-Bench run accepted no modification and therefore returned \(\mathcal{H}_0\).}
\label{tab:appendix-analysis-pool}
\end{table}

\subsubsection{Re-scoring the Recorded Decisions}
\label{sec:appendix-gate-vs-metric}

For every iteration, the controller recorded the candidate and the current confirmed harness under identical tasks and trial conditions. 
We recompute the paired difference in verification TRAIN pass rate and ask which iterations a metric-only rule (\(\Delta>0 \Rightarrow\) accept) would decide differently from the attributable-evidence gate. 
Nothing is re-executed: both rules are applied to the same stored evidence, so the comparison is exact for these runs.

\begin{figure*}[t]
\centering
\includegraphics[width=\textwidth]{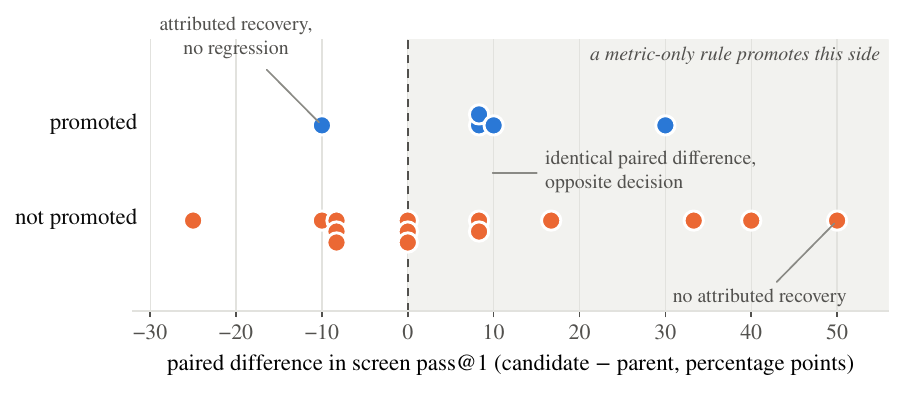}
\caption{Candidate-minus-current-harness differences in TRAIN pass rate on matched tasks and trials for 19 completed initial-verification comparisons across four OpenCode runs. Rows indicate the attributable-evidence decisions, and the shaded region marks positive differences that a metric-only rule would accept.}
\label{fig:appendix-gate-vs-metric}
\end{figure*}

The two rules differ on 7 of the 19 paired iterations (Figure~\ref{fig:appendix-gate-vs-metric}): the metric-only rule would accept 10 modifications, whereas the attributable-evidence gate accepted 5. 
Six iterations improved the verification pass rate but were not accepted. 
In the largest disagreement, a candidate gained 50.0 percentage points on its verification batch, but Trajectory Diagnosis found no task recovery attributable to a modified component. 
Conversely, one iteration advanced despite a 10.0-point verification-batch decrease because it showed an attributable recovery of the targeted behavior with no attributable regression; its final acceptance still required improvement on the confirmation batch. 
The Terminal run gives the clearest example of rejection: its only iteration with a positive paired difference gained 40.0 points without an attributable recovery, and the run therefore returned \(\mathcal{H}_0\), as reported in Table~\ref{tab:main-results}.

This re-scoring shows that the two rules select different modifications from the same within-run evidence; it does not show how the counterfactual metric-only harnesses would perform because those harnesses were not constructed in these runs. 
The separate \emph{Metric-Only Gate} ablation in Table~\ref{tab:verification-results} provides the corresponding TEST result.

\subsubsection{Which Components Were Edited}
\label{sec:appendix-channel-analysis}

We identify the native component modified in each iteration from the workspace diff captured after the editor ran, rather than from the component proposed by Experience Analysis. 
The resulting record illustrates how task properties and component visibility shape iteration, although one run per benchmark does not support a rate estimate. 
Because every run uses OpenCode, all four draw from the same component inventory; nevertheless, the components modified differ across benchmarks. 
In the BIRD run included in this decision-level analysis, all five iterations modify benchmark tool or parameter descriptions, whereas the three completed Terminal edits modify general instructions. 
Retail and Banking edits center on skills, with fewer instruction and tool-description edits. 
General instruction components are attempted most often but are seldom accepted in these runs, whereas skill and tool-schema edits do. 

\subsubsection{Where the Budget Goes}
\label{sec:appendix-budget-profile}

\begin{figure}[t]
\centering
\includegraphics[width=\columnwidth]{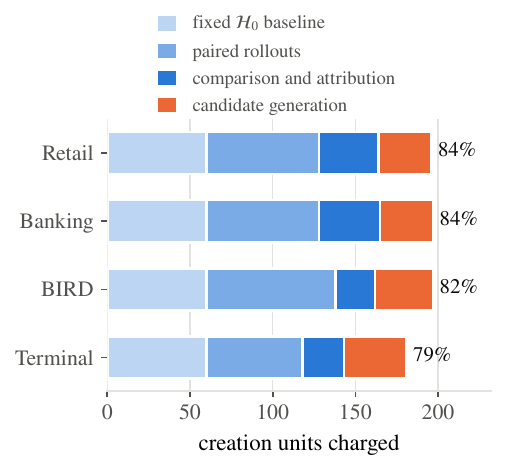}
\caption{Breakdown of interaction-budget consumption across the four OpenCode runs.
Each stacked bar shows the units spent on the initial rollout, paired verification rollouts, comparison and attribution, and candidate generation.
Percentage labels give the combined share of the first three categories.}
\label{fig:appendix-budget-profile}
\end{figure}

Aggregating the interaction-budget ledger described in \secref{sec:appendix-budget} for each run yields Figure~\ref{fig:appendix-budget-profile}. Initial and verification rollouts, behavior comparison, and post-rollout Trajectory Diagnosis account for 82.2\% of all units charged across the four runs and between 79.0\% and 83.8\% within individual runs. 
Context Exploration, proposal selection, and candidate editing account for the remainder. 
These values are direct sums from the budget ledger.
This accounting makes the comparison in Table~\ref{tab:budget} conservative because \method's 200-unit budget counts both task trials and LLM sessions, while the baseline figures count task rollouts only and exclude non-rollout LLM sessions.

\end{document}